\documentclass[letterpaper, 10 pt, conference]{ieeeconf}
\IEEEoverridecommandlockouts
\usepackage{amsmath,amsfonts}
\usepackage{algorithmic}
\usepackage{algorithm}
\usepackage{array}
\usepackage[caption=false,font=normalsize,labelfont=sf,textfont=sf]{subfig}
\usepackage{textcomp}
\usepackage{stfloats}
\usepackage{url}
\usepackage{verbatim}
\usepackage{graphicx}
\usepackage{cite}
\usepackage{multirow} 
\usepackage{pifont} 
\usepackage{colortbl} 
\usepackage{xcolor}

\usepackage{hyperref}
\hypersetup{
	colorlinks=true,
	linkcolor=cyan,
	filecolor=blue,      
	urlcolor=black,
	citecolor=green,
}

\begin{document}


\title{\LARGE \bf
TSCLIP: Robust CLIP Fine-Tuning for Worldwide Cross-Regional Traffic Sign Recognition
}


\author{Guoyang Zhao, Fulong Ma, Weiqing Qi, Chenguang Zhang, Yuxuan Liu, Ming Liu, and Jun Ma
\thanks{G. Zhao, F. Ma, W. Qi, M. Liu, and J. Ma are with the Robotics and Autonomous Systems Thrust, The Hong Kong University of Science and Technology (Guangzhou), Guangzhou, China
(e-mail: {\{gzhao492, wqiad, fmaaf\}@connect.hkust-gz.edu.cn; eelium@hkust-gz.edu.cn; jun.ma@ust.hk).} \textit{(Corresponding author: Jun Ma.)}}
\thanks{C. Zhang is with Wuhan Polytechnic University, Wuhan, China
(e-mail: qwe934063437@gmail.com).}
\thanks{Y. Liu is with the Department of Electronic and Computer Engineering, The Hong Kong University of Science and Technology, Hong Kong SAR, China
(e-mail: yliuhb@connect.ust.hk).}
}



\maketitle
\thispagestyle{empty}

\begin{abstract}


Traffic sign is a critical map feature for navigation and traffic control. Nevertheless, current methods for traffic sign recognition rely on traditional deep learning models, which typically suffer from significant performance degradation considering the variations in data distribution across different regions.
In this paper, we propose TSCLIP, a robust fine-tuning approach with the contrastive language-image pre-training (CLIP) model for worldwide cross-regional traffic sign recognition. We first curate a cross-regional traffic sign benchmark dataset by combining data from ten different sources.
Then, we propose a prompt engineering scheme tailored to the characteristics of traffic signs, which involves specific scene descriptions and corresponding rules to generate targeted text descriptions.
During the TSCLIP fine-tuning process, we implement adaptive dynamic weight ensembling (ADWE) to seamlessly incorporate outcomes from each training iteration with the zero-shot CLIP model. This approach ensures that the model retains its ability to generalize while acquiring new knowledge about traffic signs.
To the best knowledge of authors, TSCLIP is the first contrastive language-image model used for the worldwide cross-regional traffic sign recognition task.
The project website is available at: 
\href{https://github.com/guoyangzhao/TSCLIP}{https://github.com/guoyangzhao/TSCLIP}. 


\end{abstract}



\vspace{-5pt}

\section{Introduction}
Traffic sign recognition is a critical perceptual task in autonomous and assisted driving systems\cite{mogelmose2012vision}. 
Traffic signs provide rich map features and road navigation information, which are crucial for driving safety and understanding the current scene\cite{ertler2020mapillary}. Traditional traffic sign classification methods mainly rely on manually designed and extracted features such as color or shape, and use parameter-based classifiers for recognition \cite{timofte2014multi}. These methods are heavily dependent on feature parameter tuning, making them susceptible to varying scenarios, resulting in lower recognition robustness.

In recent years, convolutional neural networks (CNNs) have achieved automatic feature extraction and learning in high-dimensional spaces\cite{zhao2021real, qi2024clrkdnet}, significantly reducing the difficulty of feature design and improving recognition performance. Using deep learning methods, high accuracy results have been achieved in the field of traffic sign recognition, far surpassing traditional feature design methods\cite{almutairy2019arts}. However, since CNNs are trained only on their respective datasets, their performance significantly deteriorates when tested on datasets from different environments or regions, even for the same categories\cite{wortsman2022robust,ma2023every}. Furthermore, some datasets\cite{zhu2016traffic,tabernik2019deep,almutairy2019arts,shakhuro2016russian} use only symbols to represent categories. While this has little impact on general classification tasks, it has a substantial impact on driving tasks that rely on the semantic information of different signs to understand the environment.


\begin{figure}[t]
    \centering
    \includegraphics[width=0.5\textwidth]{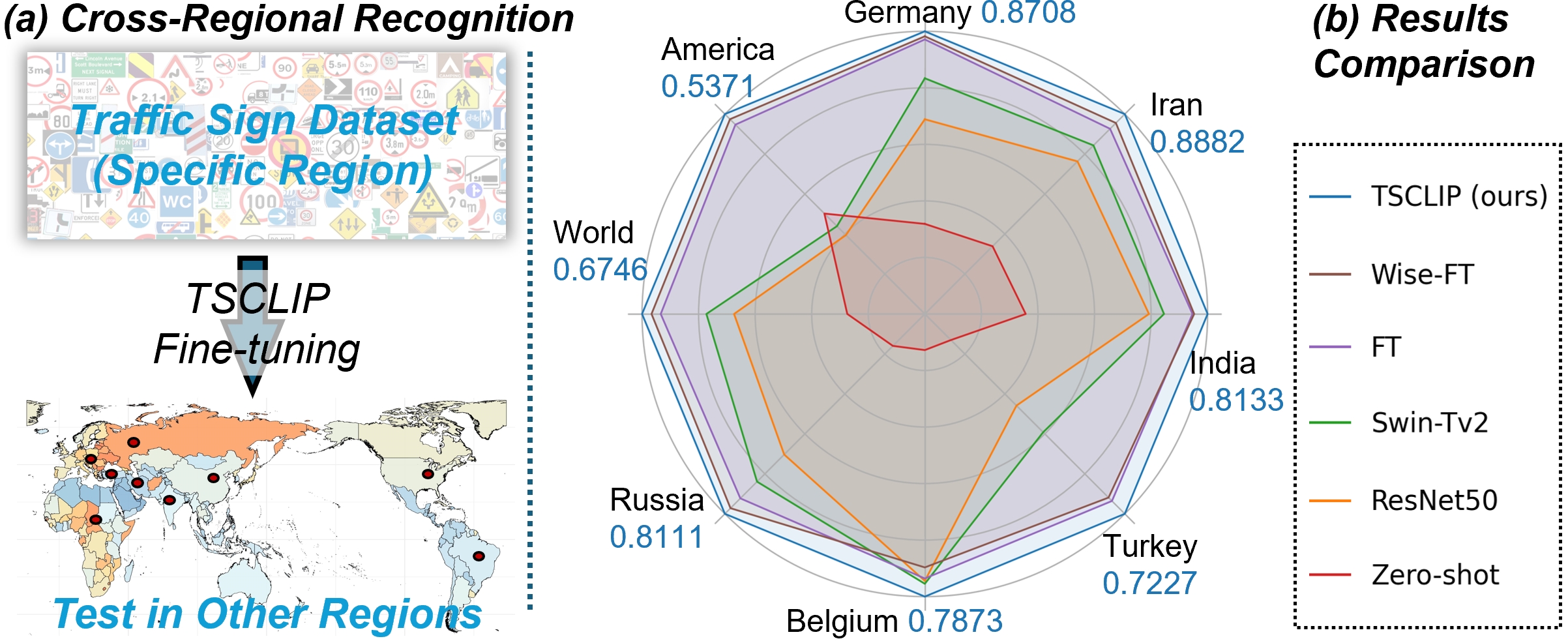}
    \vspace{-18pt}
    \caption{\textbf{Traffic sign cross-regional recognition and results.} (a) introduces the main content, fine-tuning TSCLIP on specific traffic sign datasets, and then performing recognition on other worldwide regions. (b) shows our TSCLIP model is far superior to the classic model and exceeds the mainstream scheme.}
    \label{cover-figure}
    \vspace{-18pt}
\end{figure}

The aforementioned issues have gradually been addressed with the introduction of contrastive language-image pre-training (CLIP) \cite{radford2021learning}. 
By using independent encoders to extract features from input images and texts, CLIP aligns paired features in the same feature space and employs contrastive loss to formulate the learning objective\cite{gao2024clip}. CLIP has proven to exhibit excellent zero-shot performance in visual representation, enabling the recognition and classification of new data containing unseen category images in downstream tasks, showcasing strong recognition capabilities\cite{zhou2022conditional}. In the task of traffic sign recognition, leveraging CLIP's powerful visual-language learning capability can enhance the understanding of sign semantic information, making it possible to achieve cross-region and environment recognition.

Currently, CLIP is primarily fine-tuned to enhance its application in downstream tasks\cite{guo2023calip}. The mainstream fine-tuning strategies include linear probe (LP) and end-to-end fine-tuning (FT)\cite{radford2021learning}. However, these methods tend to confine the learned weight within the distribution of the training data, significantly compromising the generalization advantage of the CLIP model. Consequently, the test performance on other data distributions is severely affected, particularly in the task of cross-regional traffic sign recognition.

In this work, to meet the requirements of cross-regional recognition, we first extract 46 mainstream and universal traffic sign categories from 10 existing datasets, creating a cross-regional traffic sign (CRTS) dataset. Based on the categories and feature distributions of traffic signs, we propose a prompt engineering scheme specifically designed for traffic signs. Regarding the TSCLIP model, we perform fine-tuning on the pre-trained CLIP model and introduce an adaptive dynamic weight ensembling (ADWE) fine-tuning scheme. Specifically, we dynamically integrate the results of each training iteration with the CLIP zero-shot model using adaptive factors. Ensuring the model maintains the generalization capabilities of the zero-shot model while learning new traffic sign knowledge.
Our primary contributions are as follows:


\begin{itemize}
\item[1)] We propose the ADWE for fine-tuning CLIP, which ensures robust cross-region recognition while effectively capturing domain-specific knowledge of traffic signs.
\item[2)] We introduce the first prompt engineering scheme tailored for traffic signs, and this significantly enhances recognition accuracy and generalization across diverse driving scenarios.
\item[3)] We establish the CRTS benchmark dataset, which serves as a robust foundation for cross-regional traffic sign testing and evaluation.
\item[4)] We develop TSCLIP, the first comparative language-image model designed for traffic sign recognition, which achieves SOTA cross-region testing performance and outperforming mainstream benchmark models.
\end{itemize}

\section{Related Works}

\subsection{Traffic Sign Recognition}
Current research on traffic sign recognition mainly falls into two categories: feature-based machine learning methods and deep learning methods for automatic feature extraction.

In feature-based machine learning methods, \cite{kerim2021recognition} and \cite{soni2019improving} employed color, histogram of oriented gradients, and local binary patterns for feature design and extraction, followed by artificial neural networks for traffic sign classification. \cite{wang2022research} and \cite{sapijaszko2019traffic} constructed frameworks based on multilayer perceptrons and support vector machines (SVM), where\cite{wang2022research} designed a logistic regression classification system, and \cite{sapijaszko2019traffic} used discrete wavelet transform and cosine transform for feature design and extraction. \cite{namyang2020thai} combined SVM and random forest algorithms, which used color descriptors for feature extraction. These methods require researchers to manually design features and classifiers, which is labor-intensive and demands specialized knowledge\cite{lim2023recent}. Moreover, the manually designed features can be biased and are not well-suited for the diverse traffic signs from different regions.

The advantage of deep learning methods lies in their ability to automatically learn and extract complex features from data, leading to higher accuracy \cite{liu2023test}. \cite{zheng2022evaluation} evaluated various CNNs and vision transformer models, showing that CNNs perform better in traffic sign classification. \cite{fu2021traffic} proposed multi-scale CNN approaches that performed well in multiple datasets. 
\cite{yazdan2021improving} proposed a novel method using a limited common image set for traffic sign recognition, demonstrating excellent accuracy. \cite{mamatkulovich2022lightweight} developed a lightweight CNN model that achieved near-perfect accuracy on the GTSRB dataset. \cite{bhatt2022real} proposed a CNN model that achieved over 91\% accuracy on an Indian dataset. Although deep learning methods can automatically learn features of different signs, their performance is often excellent only on the training scene. Once transferred to unseen scenarios, the robustness and generalization significantly decline.

\subsection{CLIP Fine-Tuning Method}


CLIP is pre-trained on a large-scale image-text dataset, which utilizes independent encoders to extract features from input images and texts, aligning these features within the same embedding space. 

\subsubsection{\textbf{Zero-Shot (ZS)}}
ZS learning is a core advantage of the CLIP model. Leveraging the alignment of image-text features learned during pre-training, CLIP can directly classify new tasks without any task-specific training data\cite{radford2021learning}. The ZS learning relies on the following formula:
\begin{equation}
\text{sim}(\mathbf{z}_i, \mathbf{t}_c) = \frac{\mathbf{z}_i \cdot \mathbf{t}_c}{\|\mathbf{z}_i\| \|\mathbf{t}_c\|}
\end{equation}
where \(\mathbf{z}_i\) is the image feature vector, \(\mathbf{t}_c\) is the text feature vector, and \(\text{sim}(\cdot)\) denotes cosine similarity. By comparing the similarity between the image and text features of each class, the highest similarity is selected as the predicted result. 



However, for specialized downstream tasks such as traffic sign recognition, fine-tuning the CLIP zero-shot model is necessary to ensure high performance. The mainstream fine-tuning methods for the pre-trained model primarily include linear probing, full finetuning, and weight ensembling.

\subsubsection{\textbf{Linear Probing (LP)}}
The LP method adds a linear classifier on top of the pre-trained model to fine-tune it. LP aims to quickly adapt to new tasks without significantly adjusting the weights of the pre-trained model. Its optimization objective is as follows:
\begin{equation}
\mathcal{L}_{\text{LP}} = \frac{1}{N} \sum_{i=1}^{N} \text{CrossEntropy}(\mathbf{W} \mathbf{z}_i, y_i)
\end{equation}
where \(\mathbf{W}\) is the weight matrix of the linear classifier, \(\mathbf{z}_i\) is the image feature, and \(y_i\) is the corresponding label. 



\subsubsection{\textbf{Full Fine-Tuning (FFT)}}
FFT updates all parameters of the pre-trained model to adapt to specific tasks. The optimization objective involves updating the weights of both the image and text encoders. The loss function defined as:
\begin{equation}
\mathcal{L}_{\text{FFT}} = \frac{1}{N} \sum_{i=1}^{N} \text{CrossEntropy}(\mathbf{W} \mathbf{z}_i, y_i) + \lambda \|\theta - \theta_0\|^2
\end{equation}
where \(\theta\) represents the model weights, \(\theta_0\) are the pre-trained model weights, and \(\lambda\) is a regularization parameter to prevent overfitting. 
FFT can be performed using the backpropagation algorithm to update all parameters, allowing the model to better adapt to new tasks.



\subsubsection{\textbf{Weight Ensembling (Wise-FT)}}
Weight Ensembling \cite{wortsman2022robust} method ensembles the weights by linearly interpolating between the weights of the zero-shot model and a fine-tuned model. The specific formula is as follows:
\begin{equation}
\theta_{\text{ensemble}} = \alpha \cdot \theta_{\text{ZS}} + (1 - \alpha) \cdot \theta_{\text{FT}}
\end{equation}
where \(\theta_{\text{ZS}}\) represents the zero-shot weight, \(\theta_{\text{FT}}\) represents the fine-tuned weight, and \(\alpha\) is the interpolation factor. This approach balances the generalization of the zero-shot model and the task-specific adaptability of the fine-tuned model.


\section{Methodology}


\subsection{Cross-Regional Traffic Sign (CRTS) Dataset }

\subsubsection{\textbf{Dataset Construction}}
The establishment of the traffic sign joint dataset from multiple regions is fundamental for cross-regional recognition tests. While some open-source traffic sign datasets \cite{timofte2014multi,tabernik2019deep,safavi2024persian} are available from different regions, their inconsistent standards, varied category counts, and differing classification criteria make them unsuitable for direct testing of model robustness across regions.

In this study, we created a CRTS joint dataset based on mainstream open-source traffic sign datasets. To ensure regional diversity, we selected datasets from 10 different countries or regions\cite{ertler2020mapillary,timofte2014multi,almutairy2019arts,zhu2016traffic,tabernik2019deep,shakhuro2016russian,houben2013detection,safavi2024persian,jodh2023indiantrafficsigns,erdem2023trafficsignturkey}. We extracted 46 commonly used categories by analyzing the distribution of similar categories and traffic sign regulations\cite{dewar2023designing} across these datasets. Every dataset was cleaned, and all categories were standardized with corresponding names.

Table~\ref{Table1} presents the parameters of the traffic sign data from 10 different regions included in the CRTS joint dataset. Due to the limitations in the creation and collection of previous datasets, not all regions include all 46 common categories. Therefore, during model training, 2 datasets are selected for joint training to ensure coverage of all categories.

\begin{table}[t!]
\renewcommand\arraystretch{1.2}
\caption{Sources of the ten regions in the CRTS joint dataset.}
\vspace{-8pt}
\centering
\footnotesize
\begin{tabular}{cccccccc}
\hline
No. & Region & Source & Category & Image & Year\\
\hline
1 & China & TT00\cite{zhu2016traffic} & 36 & 13012 & 2016  \\
2 & Germany & GTSRB\cite{houben2013detection} & 31 & 35939 & 2013  \\
3 & Iran & PTSD\cite{safavi2024persian} & 26 & 11198 & 2024  \\
4 & India & IndiaTS\cite{jodh2023indiantrafficsigns} & 41 & 3723 & 2022  \\
5 & Turkey & TurkeyTS\cite{erdem2023trafficsignturkey} & 43 & 9663 & 2020  \\
6 & Belgium & BelgiumTS\cite{timofte2014multi} & 36 & 4194 & 2014  \\
7 & Russia & RTSD\cite{shakhuro2016russian} & 44 & 56138 & 2016  \\
8 & World & MTSD\cite{ertler2020mapillary} & 45 & 37053 & 2020  \\
9 & Slovenia & DFG\cite{tabernik2019deep} & 42 & 4769 & 2019  \\
10 & America & ARTS\cite{almutairy2019arts} & 27 & 15393 & 2019  \\
\hline
\end{tabular}
\label{Table1}
\vspace{-10pt}
\end{table}

\subsubsection{\textbf{Difference of Cross-Regional Traffic Signs}}
\begin{figure}[t!]
    \centering
    \includegraphics[width=.4\textwidth]{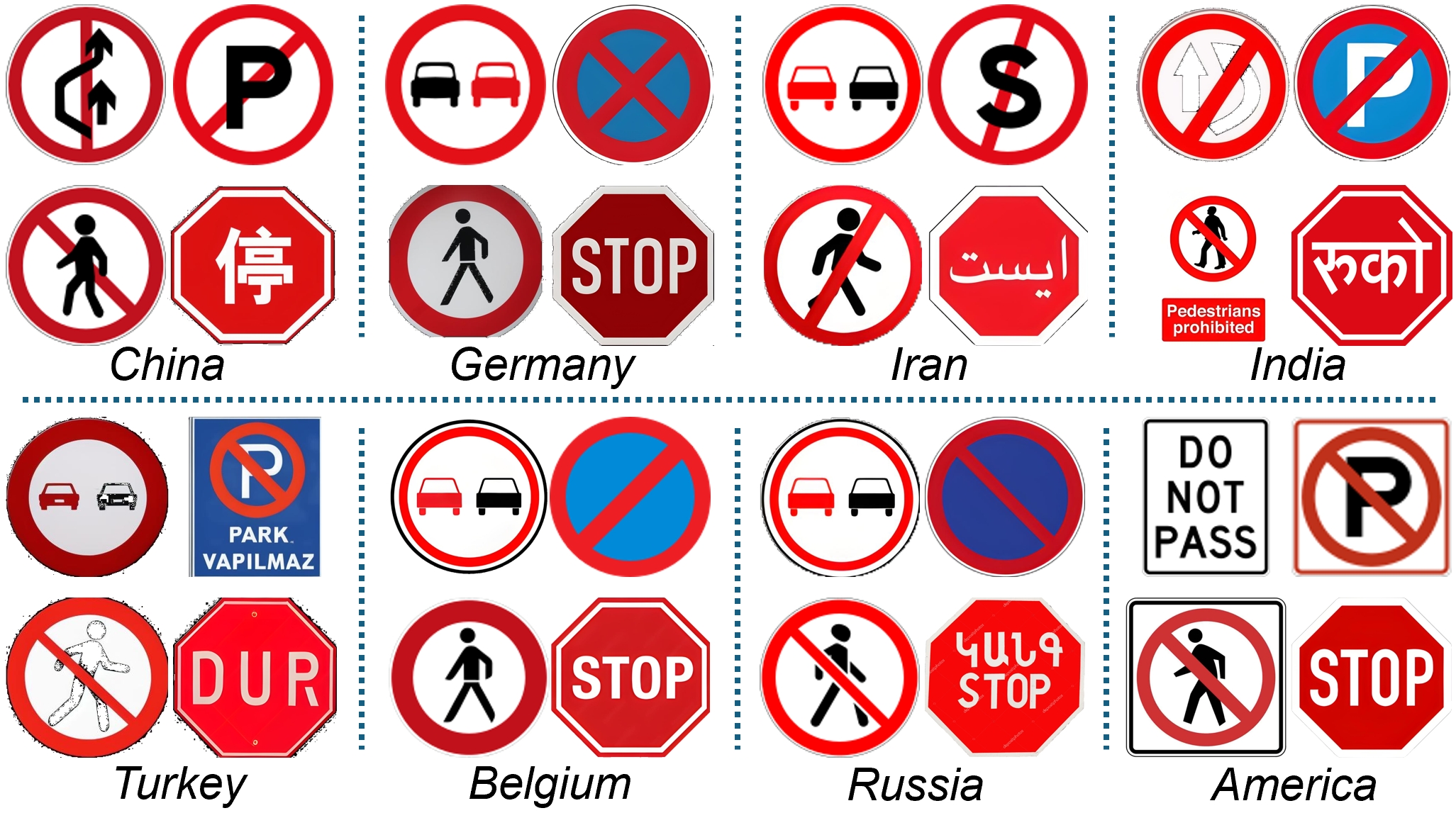}
    \vspace{-7pt}
    \caption{\textbf{Pattern differences of cross-regional samples.} Four representative traffic signs (No Overtaking, No Parking, No Pedestrians, and Stop).}
    \label{data-sample}
    \vspace{-15pt}
\end{figure}

Fig.~\ref{data-sample} shows examples of four traffic sign categories (No Overtaking, No Parking, No Pedestrians, and Stop) in different regional contexts. Traffic sign patterns vary across countries and regions, influenced by local culture and traffic regulations. Some regions have unique sign patterns, such as the No Overtaking signs in China and America. Additionally, some signs incorporate local languages, as seen in the Stop signs from China, Iran, India, Turkey, and Russia. Therefore, cross-regional traffic sign recognition poses a significant challenge, requiring models to handle continuously changing patterns.

\subsection{Traffic Sign Prompt Engineering}


To maximize the advantages of CLIP's contrastive training in both image and language modalities, we propose a prompt engineering scheme specifically designed for traffic sign classification. This scheme comprehensively considers the scene descriptions of traffic signs in real-world environments, as well as the descriptions of different categories and their corresponding traffic rules. This is the first traffic sign prompt method that provides a comprehensive description.

\subsubsection{\textbf{Structure of Prompt}}
The traffic sign prompt template is designed as a combination of two components:
\begin{equation}
\mathbf{P_i} = \mathbf{S_i} + \mathbf{T_i}
\end{equation}

where
$\mathbf{P_i}$ denotes the $i$-th prompt template, 
$\mathbf{S_i}$ represents the scenario description, and
$\mathbf{T_i}$ encompasses the traffic sign category and the associated traffic rules.

\subsubsection{\textbf{Refinement of Scenario Descriptions ($\mathbf{S_i}$)}}

The $\mathbf{S_i}$ encompasses four critical elements (a-d):

\textbf{a.} Detailed description of $\mathbf{T_i}$ categories: $\mathbf{S_{i1}} = \{s_{i1,1}, s_{i1,2}, \ldots, s_{i1,n}\}$,
where $\mathbf{S_{i1}}$ is the set of detailed descriptions for the category of $\mathbf{T_i}$, and $s_{i1,j}$ denotes each word of specific description element.

\textbf{b.} Appearance description (pattern, color, font, and shape): $\mathbf{S_{i2}} = \{s_{i2,1}, s_{i2,2}, \ldots, s_{i2,m}\}$,
where $\mathbf{S_{i2}}$ is the set of appearance descriptors for the category of $\mathbf{T_i}$, and $s_{i2,k}$ represents each word of specific appearance feature.

\textbf{c.} Background information (location and road type): $\mathbf{S_{i3}} = \{s_{i3,1}, s_{i3,2}, \ldots, s_{i3,l}\}$,
where $\mathbf{S_{i3}}$ represents the set of background information elements for the category of $\mathbf{T_i}$, and $s_{i3,l}$ denotes each word of specific background detail.

\textbf{d.} Image characteristics (resolution, quality, etc.): $\mathbf{S_{i4}} = \{s_{i4,1}, s_{i4,2}, \ldots, s_{i4,p}\}\textbf{}$,
where $\mathbf{S_{i4}}$ represents the set of image characteristics for the category of $\mathbf{T_i}$, and $s_{i4,p}$ denotes each word of specific image feature.

Thus, the scenario description $\mathbf{S_i}$ can be formulated as:
\begin{equation}
\mathbf{S_i} = \sum_{j=1}^{n} s_{i1,j} + \sum_{k=1}^{m} s_{i2,k} + \sum_{l=1}^{l} s_{i3,l} + \sum_{p=1}^{p} s_{i4,p}
\end{equation}

\subsubsection{\textbf{Traffic Sign Category and Rules ($\mathbf{T_i}$)}}
Each $\mathbf{T_i}$ in our prompt template includes the traffic sign category $\mathbf{C(T_i)}$ and the corresponding traffic rules $\mathbf{R(T_i)}$, represented as:
\begin{equation}
\mathbf{T_i} = \mathbf{C(T_i)} + \mathbf{R(T_i)}
\end{equation}
By following this structured approach, we create $n$ diverse and dynamic prompt templates $\{\mathbf{P_1}, \mathbf{P_2}, \ldots, \mathbf{P_n}\}$, ensuring a comprehensive representation of traffic signs in various scenarios. This method enhances the model's ability to understand traffic signs by providing rich contextual information and explicit traffic rules, leading to more robust recognition.




\subsection{TSCLIP Fine-Tuning Implementation}

\begin{figure*}[t!]
    \centering
    \includegraphics[width=.9\textwidth]{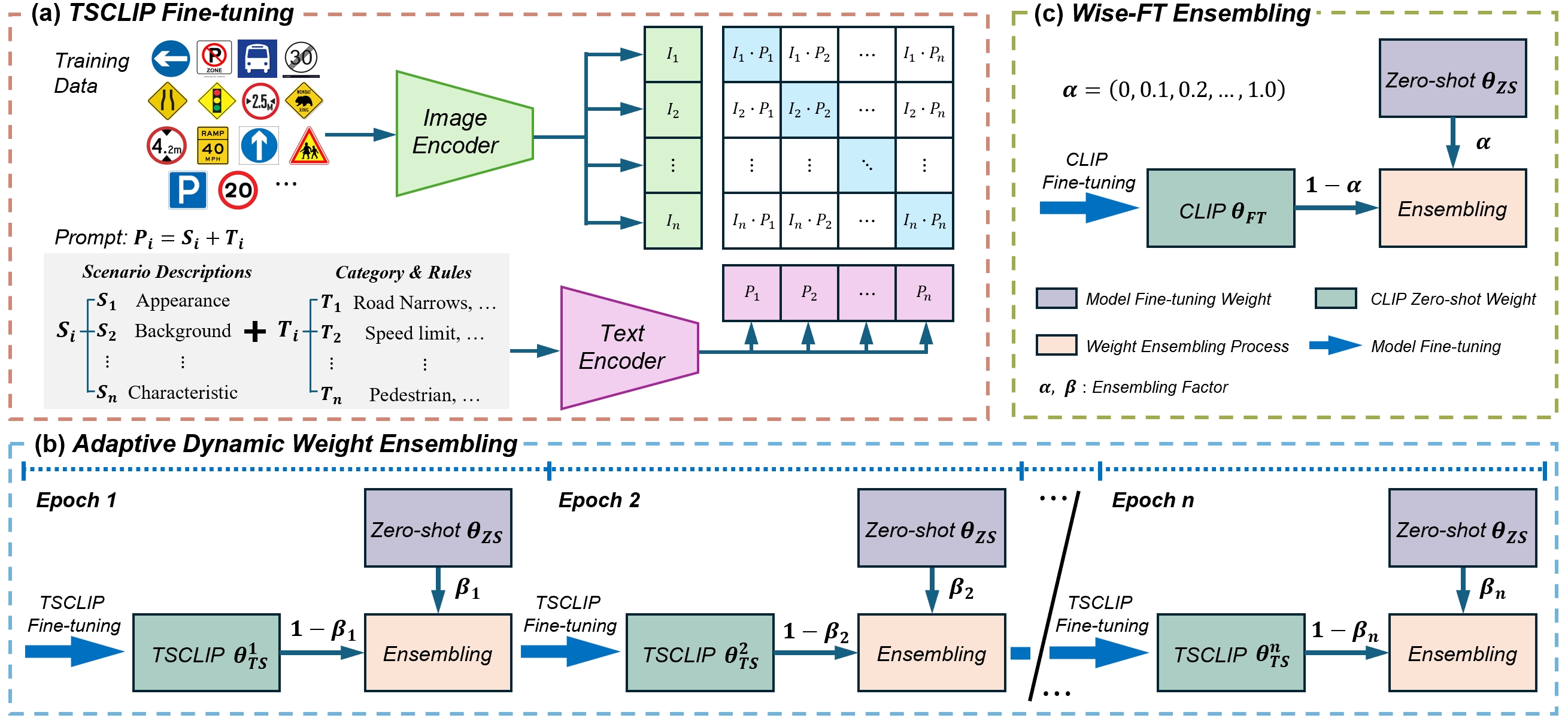}
    \vspace{-8pt}

    \caption{\textbf{Robust fine-tuning framework for TSCLIP model.} (a) shows the contrastive language-image training process of TSCLIP with traffic sign prompts.
(b) shows our proposed ADWE scheme for weight ensembling. (c) shows the Wise-FT scheme.
}
    \label{framework}
    \vspace{-15pt}
\end{figure*}

Fine-tuning a zero-shot model on a specific dataset can achieve significant performance improvements on the target distribution. However, this fine-tuning comes at the cost of robustness \cite{radford2021learning}, with the accuracy of the fine-tuned model significantly decreasing when tested on different data distributions, such as traffic signs from different regions.


The core aim of the TSCLIP model's robust fine-tuning framework (Fig.~\ref{framework}) is to combine the excellent generalization of the zero-shot model across all data distributions with the recognition capability of the fine-tuned model on traffic sign training data. This ensures the best of both worlds for specialized tasks and cross-regional generalization. Specifically, the fine-tuning framework consists of two main parts. Fig.~\ref{framework}(a) shows the initial fine-tuning of the TSCLIP using image data and corresponding prompts from the training samples, ensuring the fine-tuned model learns domain-specific knowledge of traffic signs. Fig.~\ref{framework}(b) illustrates the adaptive dynamic weight ensembling of the latest training weights with the CLIP zero-shot weight at the end of each training epoch, ensuring that the model retains a certain level of zero-shot generalization. Additionally, the robustness gains achieved by TSCLIP during the fine-tuning process do not incur extra computational costs during fine-tuning or inference.

\vspace{-5pt}
\subsection{Adaptive Dynamic Weight Ensembling (ADWE)}

\subsubsection{\textbf{Weight Ensembling}}
Interpolating model parameters is a classic idea in convex optimization\cite{polyak1990new}. Previous studies have shown that interpolation in the weight space can improve performance when models share part of the optimization trajectory \cite{neyshabur2020being}. Wise-FT is the first empirical study to explore the interpolation of non-convex models, specifically CLIP, from the perspective of distributional robustness. Fig.~\ref{framework}(c) illustrates Wise-FT's ensembling method, which involves weight ensembling with the zero-shot model after the entire fine-tuning of the CLIP is completed. This weight ensembling method enhances robustness by forcibly injecting the weights of the zero-shot model into the final training results. Although experimental results show significant improvements, this approach overlooks the dynamic nature of CLIP weights during the fine-tuning process.

\subsubsection{\textbf{Dynamic Weight}}

Our proposed dynamic weight ensembling method (Fig.~\ref{framework}(b)) involves integrating a certain proportion of the zero-shot model's parameters into the existing model weights at the end of each training epoch, followed by the next round of training. This ensures that the weights are dynamically adjusted in each epoch. This approach ensures the continuous incorporation of zero-shot knowledge throughout the fine-tuning process, thereby maintaining generalization and robustness to the greatest extent possible while learning domain-specific knowledge of the target distribution.

\subsubsection{\textbf{Adaptive Factor}}
In our dynamic weight ensembling method, since the weights obtained from each epoch are dynamically changing and the ensembled weights are further iteratively trained, a fixed mixing coefficient is not sufficient to handle the dynamic nature of the weights. The coefficient must adapt dynamically to the ever-changing training.

We propose a hybrid approach that combines cosine annealing with adaptive loss-based coefficient adjustment. Cosine annealing smoothly and non-linearly reduces the coefficient with the number of iterations, ensuring stability during the initial and final stages of training. The loss-based adjustment dynamically tunes the coefficient based on the relative performance of the training model and the zero-shot model, ensuring effective learning throughout the training process. This hybrid method balances generalization capabilities with task-specific learning needs, thereby enhancing the overall model performance across different regions.

The adaptive factor $\beta^{(t)}$ at epoch $t$ is defined as follows:
\begin{equation}
\beta^{(t)} = \left( \frac{1 + \cos\left(\frac{\pi \cdot t}{2 \cdot T}\right)}{2 \cdot \gamma} \right) \cdot \frac{L_{\text{train}}^{(t)} + L_{\text{zero-shot}}^{(t)}}{L_{\text{zero-shot}}^{(t)}}
\end{equation}
where
$T$ is the total number of epochs, $\gamma$ is a scaling factor, $L_{\text{train}}^{(t)}$ is the training loss at epoch $t$, and
$L_{\text{zero-shot}}^{(t)}$ is the zero-shot model's validation loss at epoch $t$.

The updated model weights $\theta_{\text{new}}^{(t)}$ are computed as:
\begin{equation}
\theta_{\text{new}}^{(t)} = \beta^{(t)} \cdot \theta_{\text{zero-shot}} + (1 - \beta^{(t)}) \cdot \theta_{\text{train}}^{(t)}
\end{equation}

This formula ensures that the ensemble proportion of the zero-shot model weights decreases smoothly over time while dynamically adjusting based on the relative performance of the trained model and the zero-shot model. 
This approach effectively balances the generalization and task-specific learning needs, leading to robust fine-tuning of the CLIP for traffic sign recognition across different regions.

\section{Experiment}

\subsection{Experiment Setup}


All models were trained in the PyTorch framework using the NVIDIA A100-PCIE-40GB GPUs. For training parameters, the CLIP-based models were set with a batch size of 512 and trained for 10 epochs with a learning rate of 0.00003. The classic classification model was trained based on the existing pre-trained model with a batch size of 128, trained for 100 epochs, and the learning rate was 0.0001.

\vspace{-5pt}
\subsection{Results of Cross-Regional Recognition}
We conducted comparative experiments on cross-regional traffic sign datasets, evaluating different CLIP-based methods and classical classification methods, as shown in Tables~\ref{Table2} and ~\ref{Table3}. To ensure that the training sets include all categories, we used datasets from two regions for training and the remaining eight regions for testing. Specifically, Table~\ref{Table2} shows the results from training on TT100 (China) and DFG (Slovenia) datasets, while Table~\ref{Table3} shows the results from training on RTSD (Russia) and ARTS (America) datasets.

In Table~\ref{Table2}, classical models showed unsatisfactory accuracy in cross-regional tests due to variations in traffic sign patterns across regions. Swin-T and Swin-Tv2 demonstrated the highest overall performance but still lagged significantly behind CLIP fine-tuned models. In cross-regional tests of CLIP-based models, the zero-shot model performed poorly in zero-shot classification as it had not learned such specific traffic sign categories. LP showed limited improvement as it only fine-tuned the final classifier layer without altering the model weights. 
FT and Wise-FT adjusted their weights and greatly improved the accuracy of cross-regional data, which is more than 20 percentage points higher than the classic model. Our TSCLIP model is 25 percentage points higher than the classic model and 2.5 percentage points higher than the most robust Wise-FT fine-tuning method at present.
This improvement is attributed to the integration of zero-shot model weights during the fine-tuning process, maintaining generalization and robustness while learning domain-specific knowledge of the target distribution.

\begin{table*}[t]
\renewcommand\arraystretch{1.2}
\caption{Results of Cross-Regional Recognition, Training on TT100 (China) and DFG (Slovenia) datasets.}
\vspace{-8pt}
\centering
\footnotesize
\begin{tabular}{>{\centering\arraybackslash}m{1.8cm}ccccccccccc}
\hline Difference & Methods & Germany & Iran & India & Turkey & Belgium & Russia & World & America & Avg. & $\Delta$ (\%) \\
\hline \multirow{6}{*}{Classic Model} 
& ResNet50\cite{he2016deep} & 0.5998 & 0.6781 & 0.6446 & 0.3313 & 0.7436 & 0.5705 & 0.4551 & 0.2120 & 0.5194 & - \\

& ResNet101\cite{he2016deep} & 0.5748 & 0.6539 & 0.6105 & 0.3378 & 0.7280 & 0.5748 & 0.4687 & 0.2032 & 0.5154 & -0.40\\

& EffecientNetv2 \cite{tan2021efficientnetv2} & 0.6639 & 0.7344 & 0.6857 & 0.3750 & 0.7355 & 0.6419 & 0.5115 & 0.2602 & 0.5790 & +5.96\\

& ResNext50 \cite{xie2017aggregated} & 0.6803 &  0.7313 & 0.6863 & 0.3879 & 0.7346 & 0.6809 & 0.5151 & 0.2211 & 0.5928 & +7.34\\

& Swin-T \cite{liu2021swin} & 0.7061 & 0.7426 & 0.6868 & 0.4299 & 0.7427 & 0.6991 & 0.5272 & 0.2371 & 0.6113 & +9.19\\

& Swin-Tv2 \cite{liu2022swin} & 0.7261 & 0.7491 & 0.6879 & 0.4277 & \underline{0.7516} & 0.6806 & 0.5211 & 0.2360 & 0.6086 & +8.92\\

\hline
\multirow{5}{*}{CLIP-based} 
& Zero-shot & 0.2775 & 0.3009 & 0.2901 & 0.0943 & 0.1006 & 0.1296 & 0.1850 & 0.2698 & 0.1964 & -32.30 \\

& LP \cite{radford2021learning} & 0.3056 & 0.3125 & 0.3087 & 0.1436 & 0.1397 & 0.1544 & 0.2137 & 0.2803  & 0.2222 & -29.72\\

& FT \cite{radford2021learning} & 0.8458 & 0.8229 & 0.7698 & \underline{0.6763} & 0.7368 & 0.7484 & 0.6300 & 0.5088 & 0.7230 & +20.35\\


& Wise-FT \cite{wortsman2022robust} & \underline{0.8554} & \underline{0.8487} & \underline{0.7746} & 0.6640 & 0.7060 & \underline{0.7883} & \underline{0.6520} & \underline{0.5234} & \underline{0.7442} & \underline{+22.48} \\

& TSCLIP (ours) & \textbf{0.8708} &  \textbf{0.8882} & \textbf{0.8133} & \textbf{0.7227} & \textbf{0.7873} & \textbf{0.8111} & \textbf{0.6746} & \textbf{0.5371} & \textbf{0.7695} & \textbf{+25.00} \\
\hline
\end{tabular}
\label{Table2}
\vspace{-5pt}
\end{table*}

The results of cross-regional dataset tests in Table~\ref{Table3} are generally consistent with those in Table~\ref{Table2}. 
In the cross-region test, the classic models struggle to achieve an accuracy above 0.75, while the models based on CLIP fine-tuning achieve a maximum accuracy of over 0.9.
Similar to previous findings, zero-shot and LP settings showed poor classification performance as they did not learn new weight distributions. 
Compared with the classic models, our proposed TSCLIP model continues to show the best performance, improving the accuracy by 14-16 percentage points, demonstrating the high robustness and generalization ability of our method.

\begin{table*}[t]
\renewcommand\arraystretch{1.2}
\caption{Results of Cross-Regional Recognition, Training on RTSD (Russia) and ARTS (America) datasets.}
\vspace{-8pt}
\centering
\footnotesize
\begin{tabular}{>{\centering\arraybackslash}m{1.8cm}ccccccccccc}
\hline Difference & Methods & China & Germany & Iran & India & Turkey & Belgium & World & Slovenia & Avg. & $\Delta$ (\%) \\
\hline \multirow{6}{*}{Classic Model} 
& ResNet50\cite{he2016deep} & 0.7213 & 0.6539 & 0.6523 & 0.5995 & 0.4235 & 0.6808 & 0.5682 & 0.6766 & 0.6517 & - \\

& ResNet101\cite{he2016deep} & 0.7040 & 0.6527 & 0.6441 & 0.5804 & 0.4383 & 0.6779 & 0.5571 & 0.6607 & 0.6424 &-0.93\\

& EffecientNetv2 \cite{tan2021efficientnetv2} & 0.7109 & 0.6317 & 0.6313 & 0.5896 & 0.4106 & 0.6532 & 0.5495 & 0.6464 & 0.6311 & -2.05\\

& ResNext50 \cite{xie2017aggregated} & 0.7423 &  0.6497 & 0.6467 & 0.6301 & 0.3535 & 0.6727 & 0.5936 & 0.6948 &0.6601 & +0.84\\

& Swin-T \cite{liu2021swin} & 0.7362 & 0.6887 & 0.6671 & 0.6309 & 0.4855 & 0.6932 & 0.5887 & 0.6755 &0.6754 & +2.37\\

& Swin-Tv2 \cite{liu2022swin} & 0.7500 & 0.6703 & 0.6758 & 0.6143 & 0.4732 & 0.6972 & 0.5869 & 0.6820 &0.6775 & +2.58\\

\hline
\multirow{5}{*}{CLIP-based} & Zero-shot & 0.3324 & 0.2775 & 0.3009 & 0.2901 & 0.0943 & 0.1006 & 0.1850 & 0.3085 & 0.2113 & -44.03\\

& LP \cite{radford2021learning} & 0.3426 & 0.2764 & 0.2953 & 0.3125 & 0.1179 & 0.1236 & 0.2088 & 0.3246 & 0.2291 & -42.26\\

& FT \cite{radford2021learning} & 0.9125 & 0.9027 & 0.8783 & 0.8155 & 0.7358 &  0.6986 & 0.7492 & 0.8986 & 0.7960 & +14.43 \\


& Wise-FT \cite{wortsman2022robust} & \underline{0.9251} & \underline{0.9076} & \underline{0.8802} & \underline{0.8294} & \underline{0.7481} & \underline{0.7027} & \underline{0.7576} & \underline{0.8990} & \underline{0.8032} & \underline{+15.15}\\

& TSCLIP (ours) & \textbf{0.9441} &  \textbf{0.9288} & \textbf{0.8999} & \textbf{0.8324} & \textbf{0.7620} & \textbf{0.7110} & \textbf{0.7622} & \textbf{0.9071} & \textbf{0.8138} & \textbf{+16.22}\\
\hline
\end{tabular}
\label{Table3}
\vspace{-10pt}
\end{table*}

\vspace{-5pt}
\subsection{Evaluation of Adaptive Factor}
During the fine-tuning of the TSCLIP model, each epoch's training results are adaptively dynamically weight ensembled with the zero-shot model. Therefore, the adaptive factor directly impacts the overall fine-tuning effectiveness of the model. To address this, we introduced a scaling coefficient $\gamma$ in the adaptive factor formula, which allows controlling the scale of the adaptive factor without altering the characteristics of cosine annealing. In the evaluation experiments of the adaptive factor, we set four $\gamma$ values: 1, 2, 5, and 10.

\begin{figure}[t]
    \centering
    \includegraphics[width=0.45\textwidth]{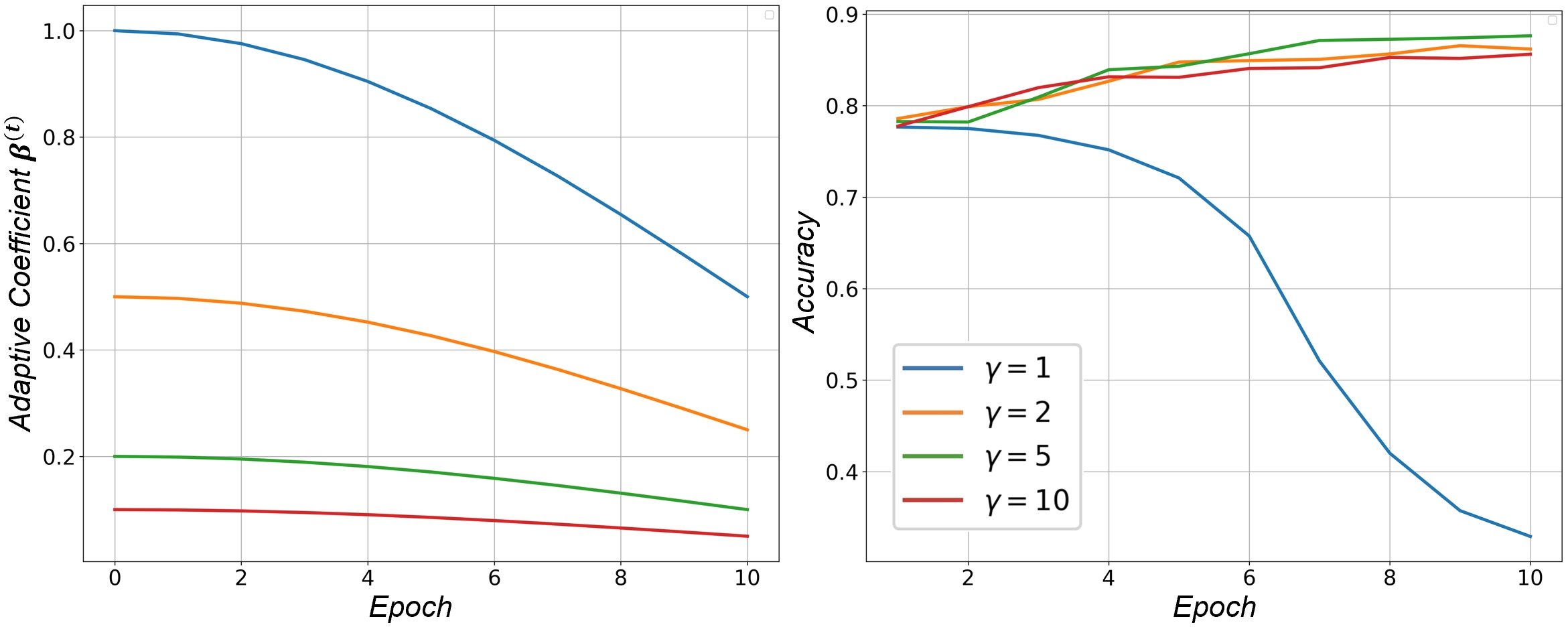}
    \vspace{-10pt}
    \caption{\textbf{Evaluation of adaptive factors.} We evaluate the fine-tuning effect of the adaptive factors under the settings of four scaling coefficient $\gamma$.}
    \label{factor}
    \vspace{-15pt}
\end{figure}

The calculation results of these four $\gamma$ values in our adaptive factor formula are shown in the left plot of Fig. \ref{factor}. The larger the $\gamma$ value, the smoother the adaptive factor $\beta^{(t)}$. The iterative training results of the TSCLIP model corresponding to these four $\gamma$ values are shown in the right plot of Fig. \ref{factor}. When $\gamma$ equals to 1, without scaling, the ensembled proportion $\beta^{(t)}$ for the zero-shot model is too high, leading to model degradation and a decrease in accuracy. When $\gamma$ equals to 2, 5, or 10, the size and variation rate of $\beta^{(t)}$ are effectively controlled, resulting in a gradual increase in overall model accuracy. The experiments showed that $\gamma$ = 5 provides the most stable ensembled proportion and the highest accuracy.

\vspace{-5pt}
\subsection{Ablation Study}
To further validate which components of the TSCLIP framework contribute most to the robustness of cross-regional traffic sign recognition, we conducted comparative ablation experiments, as shown in Table~\ref{ablation}. 
The experimental settings of the ablation study are the same as Table~\ref{Table2}.
Using Wise-FT's fine-tuning method as the baseline, we incrementally added our proposed prompt engineering scheme and the adaptive dynamic weight ensembling scheme. The results indicate that the addition of scene descriptions and rule prompts for traffic signs in the prompt engineering scheme improved accuracy by 0.8 percentage points. Furthermore, incorporating the dynamic weight training and the adaptive factor ensembling strategy increased accuracy by over 2.2 percentage points.

\begin{table}
\renewcommand\arraystretch{1.2}
\caption{Ablation study of different strategies in TSCLIP}
\vspace{-8pt}
\centering
\setlength{\tabcolsep}{2mm}
\footnotesize
\begin{tabular}{@{}cccccccc>{\columncolor{gray!20}}c@{}}
\hline
\multirow{2}{*}{Method} & \multicolumn{2}{c}{ Prompt Engineering } & & \multicolumn{2}{c}{ ADWE } & \multirow{2}{*}{Precision} & \multirow{2}{*}{$\Delta$ (\%)} \\
\cline{2-3} \cline{5-6} 
 & Scenario  & Rules & & $\theta_{\text{new}}^{(t)}$ & $\beta^{(t)}$ & & \\

\hline
Wise-FT & - & - & & - & - & 0.7375 & - \\
\hline
 \multirow{5}{*}{Ours}& \ding{51} & & & & & 0.7410 & 0.35 \\
 &  & \ding{51} & & & & 0.7412 & 0.37 \\
 & \ding{51} & \ding{51} & & & & 0.7455 & 0.80 \\
 & \ding{51} & \ding{51} & & \ding{51} & & 0.7567 & 1.92  \\
  & \ding{51} & \ding{51} & & \ding{51} & \ding{51} & 0.7683 & 3.08  \\
\hline
\end{tabular}
\label{ablation}
\vspace{-10pt}
\end{table}

\begin{figure}[t!]
    \centering
    \includegraphics[width=.5\textwidth]{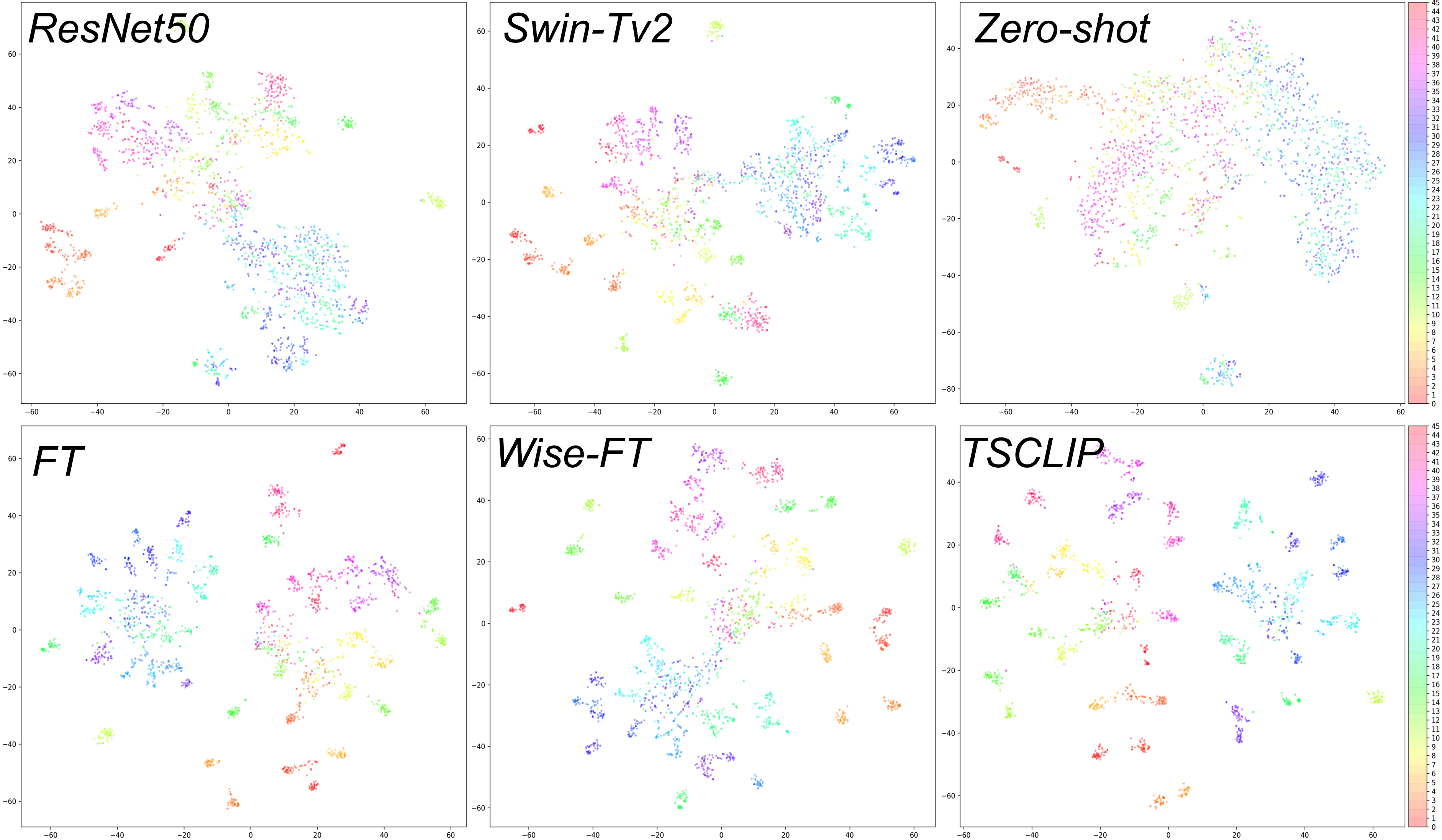}
    \vspace{-15pt}
    \caption{\textbf{T-SNE visualization of different models.} We selected two classic models and four CLIP-based models for testing on the cross-regional dataset.}
    \label{T-SNE}
    \vspace{-17pt}
\end{figure}

\vspace{-5pt}
\subsection{Model Distribution Visualization}
To better assess the cross-regional traffic sign classification capabilities of different models, we employed the t-Distributed Stochastic Neighbor Embedding (T-SNE) method to visualize high-dimensional data in two dimensions, which is a nonlinear dimensionality reduction technique.

The T-SNE visualization in Fig. \ref{T-SNE} compares two classical models and four CLIP-based models on cross-regional datasets, highlighting the challenges of adapting to diverse traffic sign patterns across regions. Classical models, like ResNet50 and Swin-Tv2, struggle with domain adaptation, as shown by the intermingling of scatter points from different categories, indicating limited clustering and generalization.

The zero-shot model performs poorly, with scatter points completely mixed, reflecting its lack of classification ability, consistent with Table \ref{Table2}. In contrast, the FT and Wise-FT fine-tuning models show improvement, with most same-category points clustering, but interwoven points remain, indicating difficulties in recognizing varied traffic sign patterns.
Our proposed TSCLIP model offers the clearest separation of categories, demonstrating superior performance in cross-regional scenarios. This improvement is attributed to the continuous integration of zero-shot weights during fine-tuning, which enhances robustness and generalization.


\section{Conclusion}
To address the challenge of robust traffic sign recognition across regions and data distributions worldwide, we proposed the TSCLIP and CRTS benchmark dataset. 
Then, we developed a prompt engineering scheme that includes specific scene descriptions and corresponding rules, specifically made for traffic sign.
The proposed ADWE method effectively combines fine-tuning model with zero-shot model, ensuring generalization to other environments while learning new traffic sign knowledge.
In extensive cross-regional tests, TSCLIP significantly outperformed mainstream benchmark methods and achieved SOTA result compared to existing robust fine-tuning methods. The ablation experiments and visual analyses further validated and illustrated the effectiveness of our approach. 
Future research will involve collecting traffic sign data worldwide to build the foundation model, enabling general recognition of traffic signs across regions.


\clearpage

{
\bibliographystyle{IEEEtran}
\bibliography{ref}

\begin{thebibliography}{10}
\providecommand{\url}[1]{#1}
\csname url@samestyle\endcsname
\providecommand{\newblock}{\relax}
\providecommand{\bibinfo}[2]{#2}
\providecommand{\BIBentrySTDinterwordspacing}{\spaceskip=0pt\relax}
\providecommand{\BIBentryALTinterwordstretchfactor}{4}
\providecommand{\BIBentryALTinterwordspacing}{\spaceskip=\fontdimen2\font plus
\BIBentryALTinterwordstretchfactor\fontdimen3\font minus \fontdimen4\font\relax}
\providecommand{\BIBforeignlanguage}[2]{{%
\expandafter\ifx\csname l@#1\endcsname\relax
\typeout{** WARNING: IEEEtran.bst: No hyphenation pattern has been}%
\typeout{** loaded for the language `#1'. Using the pattern for}%
\typeout{** the default language instead.}%
\else
\language=\csname l@#1\endcsname
\fi
#2}}
\providecommand{\BIBdecl}{\relax}
\BIBdecl

\bibitem{mogelmose2012vision}
A.~Mogelmose, M.~M. Trivedi, and T.~B. Moeslund, ``Vision-based traffic sign detection and analysis for intelligent driver assistance systems: Perspectives and survey,'' \emph{IEEE Transactions on Intelligent Transportation Systems}, vol.~13, no.~4, pp. 1484--1497, 2012.

\bibitem{ertler2020mapillary}
C.~Ertler, J.~Mislej, T.~Ollmann, L.~Porzi, G.~Neuhold, and Y.~Kuang, ``The mapillary traffic sign dataset for detection and classification on a global scale,'' in \emph{European Conference on Computer Vision}, 2020, pp. 68--84.

\bibitem{timofte2014multi}
R.~Timofte, K.~Zimmermann, and L.~Van~Gool, ``Multi-view traffic sign detection, recognition, and {3D} localisation,'' \emph{Machine Vision and Applications}, vol.~25, pp. 633--647, 2014.

\bibitem{zhao2021real}
G.~Zhao, L.~Quan, H.~Li, H.~Feng, S.~Li, S.~Zhang, and R.~Liu, ``Real-time recognition system of soybean seed full-surface defects based on deep learning,'' \emph{Computers and Electronics in Agriculture}, vol. 187, p. 106230, 2021.

\bibitem{qi2024clrkdnet}
W.~Qi, G.~Zhao, F.~Ma, L.~Zheng, J.~Ma, and M.~Liu, ``{CLRKDNet}: Speeding up lane detection with knowledge distillation,'' \emph{arXiv preprint arXiv:2405.12503}, 2024.

\bibitem{almutairy2019arts}
F.~Almutairy, T.~Alshaabi, J.~Nelson, and S.~Wshah, ``{ARTS}: Automotive repository of traffic signs for the united states,'' \emph{IEEE Transactions on Intelligent Transportation Systems}, vol.~22, no.~1, pp. 457--465, 2019.

\bibitem{wortsman2022robust}
M.~Wortsman, G.~Ilharco, J.~W. Kim, M.~Li, S.~Kornblith, R.~Roelofs, R.~G. Lopes, H.~Hajishirzi, A.~Farhadi, H.~Namkoong \emph{et~al.}, ``Robust fine-tuning of zero-shot models,'' in \emph{Proceedings of the IEEE/CVF Conference on Computer Vision and Pattern Recognition}, 2022, pp. 7959--7971.

\bibitem{ma2023every}
F.~Ma, X.~Yan, Y.~Liu, and M.~Liu, ``Every dataset counts: Scaling up monocular {3D} object detection with joint datasets training,'' \emph{arXiv preprint arXiv:2310.00920}, 2023.

\bibitem{zhu2016traffic}
Z.~Zhu, D.~Liang, S.~Zhang, X.~Huang, B.~Li, and S.~Hu, ``Traffic-sign detection and classification in the wild,'' in \emph{Proceedings of the IEEE Conference on Computer Vision and Pattern Recognition}, 2016, pp. 2110--2118.

\bibitem{tabernik2019deep}
D.~Tabernik and D.~Sko{\v{c}}aj, ``Deep learning for large-scale traffic-sign detection and recognition,'' \emph{IEEE Transactions on Intelligent Transportation Systems}, vol.~21, no.~4, pp. 1427--1440, 2019.

\bibitem{shakhuro2016russian}
V.~I. Shakhuro and A.~Konouchine, ``Russian traffic sign images dataset,'' \emph{Computer Optics}, vol.~40, no.~2, pp. 294--300, 2016.

\bibitem{radford2021learning}
A.~Radford, J.~W. Kim, C.~Hallacy, A.~Ramesh, G.~Goh, S.~Agarwal, G.~Sastry, A.~Askell, P.~Mishkin, J.~Clark \emph{et~al.}, ``Learning transferable visual models from natural language supervision,'' in \emph{International Conference on Machine Learning}, 2021, pp. 8748--8763.

\bibitem{gao2024clip}
P.~Gao, S.~Geng, R.~Zhang, T.~Ma, R.~Fang, Y.~Zhang, H.~Li, and Y.~Qiao, ``{CLIP-Adapter}: Better vision-language models with feature adapters,'' \emph{International Journal of Computer Vision}, vol. 132, no.~2, pp. 581--595, 2024.

\bibitem{zhou2022conditional}
K.~Zhou, J.~Yang, C.~C. Loy, and Z.~Liu, ``Conditional prompt learning for vision-language models,'' in \emph{Proceedings of the IEEE/CVF Conference on Computer Vision and Pattern Recognition}, 2022, pp. 16\,816--16\,825.

\bibitem{guo2023calip}
Z.~Guo, R.~Zhang, L.~Qiu, X.~Ma, X.~Miao, X.~He, and B.~Cui, ``{CALIP}: Zero-shot enhancement of clip with parameter-free attention,'' in \emph{Proceedings of the AAAI Conference on Artificial Intelligence}, vol.~37, no.~1, 2023, pp. 746--754.

\bibitem{kerim2021recognition}
A.~Kerim and M.~{\"O}. Efe, ``Recognition of traffic signs with artificial neural networks: A novel dataset and algorithm,'' in \emph{2021 International Conference on Artificial Intelligence in Information and Communication}, 2021, pp. 171--176.

\bibitem{soni2019improving}
D.~Soni, R.~K. Chaurasiya, and S.~Agrawal, ``Improving the classification accuracy of accurate traffic sign detection and recognition system using hog and lbp features and pca-based dimension reduction,'' in \emph{Proceedings of International Conference on Sustainable Computing in Science, Technology and Management, Amity University Rajasthan, Jaipur-India}, 2019.

\bibitem{wang2022research}
B.~Wang, ``Research on the optimal machine learning classifier for traffic signs,'' in \emph{SHS Web of Conferences}, vol. 144, 2022, p. 03014.

\bibitem{sapijaszko2019traffic}
G.~Sapijaszko, T.~Alobaidi, and W.~B. Mikhael, ``Traffic sign recognition based on multilayer perceptron using {DWT} and {DCT},'' in \emph{2019 IEEE 62nd International Midwest Symposium on Circuits and Systems}, 2019, pp. 440--443.

\bibitem{namyang2020thai}
N.~Namyang and S.~Phimoltares, ``Thai traffic sign classification and recognition system based on histogram of gradients, color layout descriptor, and normalized correlation coefficient,'' in \emph{2020-5th International Conference on Information Technology}, 2020, pp. 270--275.

\bibitem{lim2023recent}
X.~R. Lim, C.~P. Lee, K.~M. Lim, T.~S. Ong, A.~Alqahtani, and M.~Ali, ``Recent advances in traffic sign recognition: approaches and datasets,'' \emph{Sensors}, vol.~23, no.~10, p. 4674, 2023.

\bibitem{liu2023test}
Y.~Liu, W.~Zhang, G.~Zhao, J.~Zhu, A.~V. Vasilakos, and L.~Wang, ``Test-time adaptation for nighttime color-thermal semantic segmentation,'' \emph{IEEE Transactions on Artificial Intelligence}, 2023.

\bibitem{zheng2022evaluation}
Y.~Zheng and W.~Jiang, ``Evaluation of vision transformers for traffic sign classification,'' \emph{Wireless Communications and Mobile Computing}, vol. 2022, no.~1, p. 3041117, 2022.

\bibitem{fu2021traffic}
H.~Fu and H.~Wang, ``Traffic sign classification based on prototypes,'' in \emph{2021 16th International Conference on Intelligent Systems and Knowledge Engineering}, 2021, pp. 7--10.

\bibitem{yazdan2021improving}
R.~Yazdan and M.~Varshosaz, ``Improving traffic sign recognition results in urban areas by overcoming the impact of scale and rotation,'' \emph{ISPRS Journal of Photogrammetry and Remote Sensing}, vol. 171, pp. 18--35, 2021.

\bibitem{mamatkulovich2022lightweight}
B.~B. Mamatkulovich, ``Lightweight residual layers based convolutional neural networks for traffic sign recognition,'' \emph{European International Journal of Multidisciplinary Research and Management Studies}, vol.~2, no.~05, pp. 88--94, 2022.

\bibitem{bhatt2022real}
N.~Bhatt, P.~Laldas, and V.~B. Lobo, ``A real-time traffic sign detection and recognition system on hybrid dataset using {CNN},'' in \emph{2022 7th International Conference on Communication and Electronics Systems}, 2022, pp. 1354--1358.

\bibitem{safavi2024persian}
S.~M. Safavi, H.~Seyedarabi, and R.~Afrouzian, ``Persian traffic sign classification using convolutional neural network and transfer learning,'' \emph{Arabian Journal for Science and Engineering}, pp. 1--10, 2024.

\bibitem{houben2013detection}
S.~Houben, J.~Stallkamp, J.~Salmen, M.~Schlipsing, and C.~Igel, ``Detection of traffic signs in real-world images: The german traffic sign detection benchmark,'' in \emph{The 2013 International Joint Conference on Neural Networks}, 2013, pp. 1--8.

\bibitem{jodh2023indiantrafficsigns}
\BIBentryALTinterwordspacing
S.~D. Jodh, ``Indian traffic signs prediction - 85 classes,'' 2022. [Online]. Available: \url{https://www.kaggle.com/datasets/sarangdilipjodh/indian-traffic-signs-prediction85-classes}
\BIBentrySTDinterwordspacing

\bibitem{erdem2023trafficsignturkey}
\BIBentryALTinterwordspacing
E.~Cem, ``Traffic sign images from turkey,'' 2020. [Online]. Available: \url{https://www.kaggle.com/datasets/erdicem/traffic-sign-images-from-turkey}
\BIBentrySTDinterwordspacing

\bibitem{dewar2023designing}
R.~Dewar and M.~Pronin, ``Designing road sign symbols,'' \emph{Transportation Research Part F: Traffic Psychology and Behaviour}, vol.~94, pp. 466--491, 2023.

\bibitem{polyak1990new}
B.~T. Polyak, ``A new method of stochastic approximation type,'' \emph{Avtomatika i Telemekhanika}, no.~7, pp. 98--107, 1990.

\bibitem{neyshabur2020being}
B.~Neyshabur, H.~Sedghi, and C.~Zhang, ``What is being transferred in transfer learning?'' \emph{Advances in Neural Information Processing Systems}, vol.~33, pp. 512--523, 2020.

\bibitem{he2016deep}
K.~He, X.~Zhang, S.~Ren, and J.~Sun, ``Deep residual learning for image recognition,'' in \emph{Proceedings of the IEEE Conference on Computer Vision and Pattern Recognition}, 2016, pp. 770--778.

\bibitem{tan2021efficientnetv2}
M.~Tan and Q.~Le, ``{EfficientNetV2}: Smaller models and faster training,'' in \emph{International Conference on Machine Learning}, 2021, pp. 10\,096--10\,106.

\bibitem{xie2017aggregated}
S.~Xie, R.~Girshick, P.~Doll{\'a}r, Z.~Tu, and K.~He, ``Aggregated residual transformations for deep neural networks,'' in \emph{Proceedings of the IEEE Conference on Computer Vision and Pattern Recognition}, 2017, pp. 1492--1500.

\bibitem{liu2021swin}
Z.~Liu, Y.~Lin, Y.~Cao, H.~Hu, Y.~Wei, Z.~Zhang, S.~Lin, and B.~Guo, ``{Swin Transformer}: Hierarchical vision transformer using shifted windows,'' in \emph{Proceedings of the IEEE/CVF International Conference on Computer Vision}, 2021, pp. 10\,012--10\,022.

\bibitem{liu2022swin}
Z.~Liu, H.~Hu, Y.~Lin, Z.~Yao, Z.~Xie, Y.~Wei, J.~Ning, Y.~Cao, Z.~Zhang, L.~Dong \emph{et~al.}, ``{Swin Transformer V2}: Scaling up capacity and resolution,'' in \emph{Proceedings of the IEEE/CVF Conference on Computer Vision and Pattern Recognition}, 2022, pp. 12\,009--12\,019.

\end{thebibliography}
}


 




\vfill

\end{document}